\documentclass[lettersize,journal]{IEEEtran}
\usepackage{amsmath,amsfonts}
\usepackage{algorithmic}
\usepackage{algorithm}
\usepackage{array}
\usepackage[caption=false,font=normalsize,labelfont=sf,textfont=sf]{subfig}
\usepackage{textcomp}
\usepackage{stfloats}
\usepackage{url}
\usepackage{verbatim}
\usepackage{graphicx}
\usepackage{cite}
\usepackage{hyperref}
\usepackage{color,xcolor}
\usepackage{multirow}
\usepackage{booktabs}
\usepackage{times}
\usepackage{latexsym}
\usepackage[utf8]{inputenc}
\usepackage{microtype}
\usepackage{inconsolata}
\usepackage{multirow}
\usepackage{booktabs}
\usepackage{tabularx}
\usepackage{makecell}
\usepackage{enumitem}
\usepackage{array}
\usepackage{graphicx}      % 插图支持
\usepackage{subcaption}    % 子图环境（定义了 subfigure）
\usepackage{graphicx}
\usepackage{caption}
\usepackage{subcaption}

\usepackage{booktabs, multirow, tabularx, makecell, array}
\newcommand{\figref}[1]{Figure \ref{#1}}
\newcommand{\tbref}[1]{Table \ref{#1}}
\renewcommand{\eqref}[1]{Equation \ref{#1}}

\begin{document}

\title{Plan of Knowledge: Retrieval-Augmented Large Language Models for Temporal Knowledge Graph Question Answering}

\author{Xinying Qian, Ying Zhang, Yu Zhao, Baohang Zhou, Xuhui Sui and Xiaojie Yuan
\thanks{This work was supported by the National Natural Science Foundation of China (No. 62272250), the Natural Science Foundation of Tianjin, China (No. 22JCJQJC00150). (Corresponding author: Ying Zhang.)}%
\thanks{Xinying Qian, Ying Zhang, Yu Zhao, Xuhui Sui and Xiaojie Yuan are with the Tianjin Key Laboratory of Network and Data Security Technology, College of Computer Science, Nankai University, Tianjin 300350, China, Baohang Zhou is with School of Software, Tiangong University, 300380, Tianjin, China (e-mail: qianxinying@dbis.nankai.edu.cn; yingzhang@nankai.edu.cn; zhaoyu@dbis.nankai.edu.cn; zhoubaohang@tiangong.edu.cn; suixuhui@dbis.nankai.edu.cn; yuanxj@nankai.edu.cn}}

\maketitle
\begin{abstract}
Temporal Knowledge Graph Question Answering (TKGQA) aims to answer time-sensitive questions by leveraging factual information from Temporal Knowledge Graphs (TKGs).
While previous studies have employed pre-trained TKG embeddings or graph neural networks to inject temporal knowledge, they fail to fully understand the complex semantic information of time constraints.
Recently, Large Language Models (LLMs) have shown remarkable progress, benefiting from their strong semantic understanding and reasoning generalization capabilities.
However, their temporal reasoning ability remains limited. LLMs frequently suffer from  hallucination and a lack of knowledge.
To address these limitations, we propose the Plan of Knowledge framework with a contrastive temporal retriever, which is named PoK.
Specifically, the proposed Plan of Knowledge module decomposes a complex temporal question into a sequence of sub-objectives from the pre-defined tools, serving as intermediate guidance for reasoning exploration.
In parallel, we construct a Temporal Knowledge Store (TKS) with a contrastive retrieval framework, enabling the model to selectively retrieve semantically and temporally aligned facts from TKGs.
% We also add task prompts in questions so as to make the model pay more attention to temporal constraints.
%
By combining structured planning with temporal knowledge retrieval, PoK effectively enhances the interpretability and factual consistency of temporal reasoning.
Extensive experiments on four benchmark TKGQA datasets demonstrate that PoK significantly improves the retrieval precision and reasoning accuracy of LLMs,
surpassing the performance of the state-of-the-art TKGQA methods by 56.0\% at most.

\end{abstract}

\section{Introduction}
\begin{figure}[!t]
    \centering
    \includegraphics[width=0.95\linewidth]{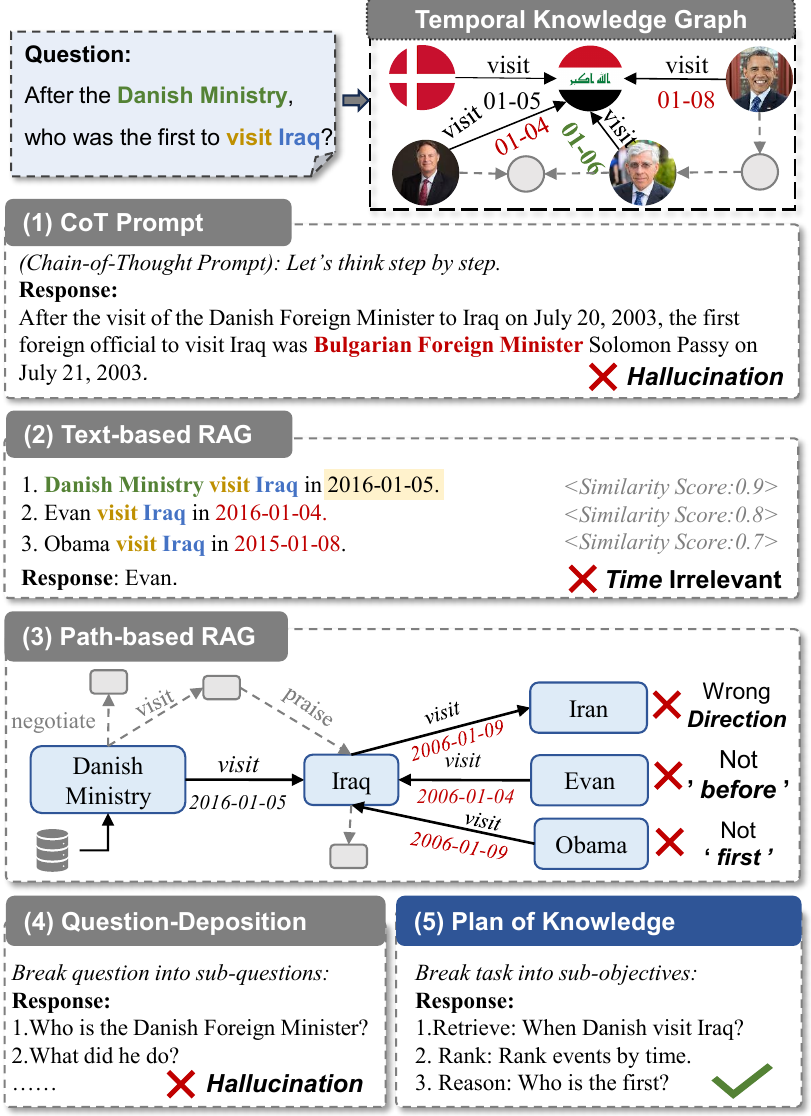}  % 用linewidth更精准
    \caption{Illustration of the key differences and challenges in integrating temporal knowledge into large language models across (1) CoT prompting, (2) text-based RAG, (3) path-based RAG, (4) question-decomposition RAG, and (5) our proposed plan of knowledge RAG framework.
    }
    \label{intro}
\end{figure}
Knowledge graphs (KGs) such as DBpedia~\cite{dbpedia}, Freebase~\cite{freebase}, and Wikidata~\cite{wikidata} organize structured knowledge in the form of triples \textit{(subject, predicate, object)}.
They have been widely adopted in various downstream applications, including question answering~\cite{sun2019pullnet}, recommender systems~\cite{guo2020survey}, and information retrieval~\cite{Zhu_2025}.
However, many real-world facts are inherently dynamic and evolve over time.
For instance, the triple \textit{(Lionel Messi, member of, FC Barcelona)} became invalid after Messi left FC Barcelona in 2021.
To capture such temporal dynamics, temporal knowledge graphs (TKGs) have been introduced, where facts are represented as quadruples \textit{(subject, predicate, object, timestamp)}.
TKGs such as ICEWS~\cite{icews} and GDELT~\cite{gdelt} continuously record large-scale political, economic, and social events with explicit temporal annotations.
These temporal extensions enable time-aware reasoning and support applications in which temporal context plays a crucial role.

Temporal knowledge graph question answering (TKGQA) aims to answer time-sensitive questions by leveraging the knowledge stored in TKGs~\cite{saxena2021question}.
Compared with conventional question answering, time-sensitive question answering poses additional challenges, as it requires not only reasoning over entities and relations but also sophisticated temporal inference across evolving events.
Traditional approaches~\cite{saxena2021question,mavromatis2022tempoqr} cast TKGQA as a temporal KG completion problem, estimating answer likelihoods via scoring functions. 
Nevertheless, these methods often struggle to fully capture the complex semantic and temporal constraints expressed in natural language questions~\cite{chen2023multi}.
In contrast, large language models (LLMs) have demonstrated substantial potential in question answering and information retrieval tasks~\cite{sun2024thinkongraph,luo2024reasoning}, effectively integrating semantic understanding with knowledge graph reasoning~\cite{huang2023reasoning,wei2023chainofthought}.
This motivates the integration of temporal knowledge into LLMs to enhance their capacity for addressing complex, multi-hop temporal questions.
However, augmenting LLMs with  temporal reasoning capabilities presents several challenges:

\textbf{(1) Hallucination in multi-hop time-sensitive questions.}
For time-sensitive questions in the TKGQA task, many questions require multi-hop reasoning, such as “After the Danish Ministry, who was the first to visit Iraq?"
Reasoning over such questions is particularly challenging, as LLMs must not only integrate information from multiple hops but also accurately capture temporal constraints expressed by keywords such as “first," “after," or “before."
As illustrated in \figref{intro}, LLMs often hallucinate when dealing with such queries, generating logically inconsistent or factually incorrect reasoning chains. Even when following a step-by-step reasoning paradigm (\figref{intro} (1)), hallucinations frequently occur in intermediate steps, leading to erroneous final answers.
Moreover, as shown in \figref{intro} (4), question decomposition methods may also suffer from hallucinations, where the model incorrectly splits a complex question into sub-questions that are completely irrelevant to the original question.
In contrast, as illustrated in \figref{intro} (5), explicitly planning a complex temporal question into a well-structured sequence of sub-tasks enables LLMs to reason more faithfully and derive more accurate conclusions. Therefore, we argue that explicit \textit{question planning of knowledge} is essential for mitigating hallucinations and enhancing the reliability of temporal reasoning.

\textbf{(2) Lack of Temporal Knowledge.}
Enhancing the temporal reasoning capability of LLMs requires access to relevant knowledge from Temporal Knowledge Graphs (TKGs).
Text-based Retrieval-Augmented Generation (RAG) methods~\cite{li2023fewshot} often leverage off-the-shelf retrievers such as BGE~\cite{chen2024bgem3embeddingmultilingualmultifunctionality} to extract supporting facts from a knowledge graph as background knowledge.
However, as in \figref{intro} (2), these approaches primarily rely on semantic similarity while overlooking the temporal constraints specified in the question. Consequently, the retrieved knowledge may be temporally inconsistent with the question.
Path-based RAG methods~\cite{sun2024thinkongraph} perform reasoning directly on the graph. Yet, TKGs introduce an additional temporal dimension, making them substantially more complex and often unable to retrieve temporally relevant facts, as in \figref{intro} (3).
Therefore, designing a retriever that jointly accounts for both semantic relevance and temporal consistency is crucial for TKGQA task.

To address these challenges, we propose PoK, a Time-aware Plan of Knowledge framework for TKGQA.
LLMs often hallucinate when facing implicit multi-hop temporal questions. To mitigate this, PoK introduces a Plan of Knowledge strategy that decomposes complex questions into structured sub-objectives, guiding stepwise exploration and integration of temporal knowledge from TKGs.
To solve the problem of lack knowledge, a temporal retrieval module further enhances reasoning by unifying semantic and temporal information.
We inject task prompts into question representations, enabling perception of temporal constraints.
Through contrastive time-aware fine-tuning, question embeddings are aligned with factual representations across both semantic and temporal dimensions, forming a structured temporal knowledge store (TKS) for efficient retrieval.
During inference, PoK performs temporal retrieval and re-ranking to ensure that selected facts are contextually relevant and temporally consistent.
Finally, retrieved evidence is fed into a fine-tuned LLM for reasoning, enabling accurate multi-hop temporal inference.
Experiments on four benchmark datasets show that PoK substantially improves temporal reasoning, outperforming baselines in both retrieval precision and answer accuracy through its integration of structured planning, temporal retrieval, and knowledge-enhanced LLM reasoning.
Overall, our work makes the following contributions:
\begin{itemize}
     \item To mitigate hallucination, we integrate LLMs with temporal knowledge and propose PoK, a Plan of Knowledge framework that incorporates pre-defined temporal operators to guide reasoning over temporal knowledge graphs. 
     \item We design a prompt-based contrastive time-aware retrieval strategy that simultaneously pays attention to semantic similarity and temporal constraints.
     \item We conduct experiments on four benchmark TKGQA datasets, demonstrating that our approach significantly improves the retrieval precision and reasoning accuracy of LLMs, achieving relative improvements of 1.8\%, 6.5\%, 54.2\%, and 56.0\% compared with the state-of-the-art TKGQA methods.
\end{itemize}

This article extends our conference version \cite{timer4} with the following improvements:
\begin{itemize}
    \item Considering that the previous framework cannot handle multi-hop reasoning, we propose a novel Plan of Knowledge framework.
    \item We further enhance the temporal retriever with LLM-based semantic representations, task prompt and an InfoNCE objective, improving both semantic discrimination and temporal sensitivity.
    \item We expand the related work with a comprehensive review of TKGQA and knowledge-enhanced LLM reasoning, including recent LLM advances, fine-tuning strategies, and RAG paradigms.
    \item We introduce more datasets and conduct extensive experiments on four datasets, providing deeper analyses of model performance.
\end{itemize}

The remaining sections of this article are organized as follows. Section \ref{relatedwork} provides a concise review of related work. Section \ref{Preliminaries} introduces the problem formulation and notations. Section \ref{Method} details our proposed framework. Section \ref{Experiments} presents the experimental setup, main results and additional analysis. Finally, Section \ref{conclusion} concludes the paper and discusses potential directions for future work.

\section{Related Work} \label{relatedwork}
\subsection{Temporal Knowledge Graph Question Answering}

Temporal Knowledge Graph Question Answering (TKGQA) requires reasoning over both entities and timestamps, making it more challenging than conventional KGQA.
Early studies formulate it as a temporal completion task, employing scoring functions to align questions with candidate facts~\cite{saxena2021question}.
Subsequent models such as TempoQR~\cite{mavromatis2022tempoqr} and MultiQA~\cite{chen2023multi} enhance temporal representation learning through contextual and multi-granularity temporal encoding.
Specifically, TempoQR integrates contextual, entity-level, and timestamp-level features via three dedicated modules, while MultiQA employs Transformer-based encoders to aggregate temporal signals across multiple granularities.

Other methods explicitly capture temporal dependencies through graph-based architectures.
EXAQT~\cite{jia2021complex} combines Relational Graph Convolutional Networks with dictionary-based matching to model relational patterns, whereas TwiRGCN~\cite{sharma2022twirgcn} introduces temporally weighted graph convolutions with answer gating to highlight relevant temporal edges.
LGQA~\cite{liu2023local} adopts a multi-hop message passing GNN to jointly encode local and global temporal contexts, and TMA~\cite{liu2023timeawaremultiwayadaptivefusion} employs a time-aware multiway adaptive fusion network to generate temporal-specific representations.

To address these limitations, recent works explore integrating LLMs with temporal reasoning.
ARI~\cite{chen2023temporal} outlines a framework for enhancing LLMs' temporal adaptability but remains limited to large-scale models.
GenTKGQA~\cite{gentkgqa} leverages LLMs to retrieve relevant temporal subgraphs and encodes them via a pre-trained temporal GNN.
TempAgent~\cite{timeagent} introduces an LLM-based autonomous agent to strengthen temporal reasoning and comprehension.
M3TQA~\cite{m3tqa} enables asynchronous alignment and fusion of PLM and GNN features through a multi-stage aggregation module.
Finally, RTQA~\cite{rtqa} decomposes questions into sub-problems solved in a bottom-up manner with LLMs and TKGs, though it still relies on manually crafted prompts and remains error-prone in the decomposition process.

\subsection{Knowledge-Augmented Large Language Models}

Large decoder-only models such as GPT-5 have driven rapid progress in NLP with strong generative and reasoning abilities. To further enhance reasoning, methods like Chain-of-Thought (CoT)~\cite{cot} and self-refine~\cite{selfrefine} decompose or iteratively improve model outputs.
Open-source models such as Llama, ChatGLM~\cite{glm2024chatglmfamilylargelanguage}, QwenChat~\cite{qwen3}, along with specialized variants like the Qwen embedding models~\cite{zhang2025qwen3embeddingadvancingtext} provide flexible foundations for downstream tasks. To adapt LLMs for specific domains, supervised finetuning (SFT) techniques based on instruction tuning have become widely adopted. Representative methods such as LoRA~\cite{lora}, QLoRA~\cite{qlora}, and P-Tuning v2~\cite{liu2022ptuningv2prompttuning} further enhance the adaptability of LLMs.

Retrieval-Augmented Generation (RAG) mitigates hallucination by integrating external knowledge. Naive RAG~\cite{chen2023benchmarkinglargelanguagemodels} follows a “retrieve-then-read" pipeline, while ReAct RAG~\cite{yao2023reactsynergizingreasoningacting} enhances interpretability through reasoning traces, achieving strong results in multi-hop QA.

Knowledge Graph Question Answering (KGQA)~\cite{9960856} incorporates structured knowledge to further reduce hallucination. 
Retrieval-based methods \cite{baek2023knowledge, he2024gretriever} focus on retrieving relevant subgraphs or triples from the KGs. The LLMs are then used to process and reason over the retrieved information.
However, they only consider semantic similarity and neglect temporal constraints.
Path-based methods \cite{9895199,cheng2024call,planongraph,long2025epermevidencepathenhanced} involve exploring paths within the KGs to establish connections between the question and the answers. These methods typically utilize LLMs to traverse the graph and generate possible paths.
However, they cannot be directly applied to TKGQA because they do not account for temporal dimension in path reasoning.
Agent-based methods\cite{sun2023think, jiang2023structgpt} treat LLMs as an agent to search and prune on the KGs to find answers.
However, they prove inefficient for complex reasoning tasks due to their reliance on multiple LLM-calls. Furthermore, the greedy decision-making process is susceptible to error propagation.

\section{Preliminaries} \label{Preliminaries}

\textbf{Temporal Knowledge Graph (TKG)} is defined as $\mathcal{G} = \{\mathcal{E}, \mathcal{P}, \mathcal{T}, \mathcal{F}\}$, 
where $\mathcal{E}$, $\mathcal{P}$, and $\mathcal{T}$ denote the sets of entities, predicates, and timestamps, respectively. 
Each temporal fact is a quadruple $(s, p, o, t) \in \mathcal{E} \times \mathcal{P} \times \mathcal{E} \times \mathcal{T}$, 
with $s$ and $o$ as subject and object, $p$ as the relation, and $t$ as the time of validity. 
The set of all facts is $\mathcal{F}$. Unlike static KGs, TKGs capture temporal dynamics, enabling reasoning over time-dependent knowledge.  

\textbf{Temporal Knowledge Graph Question Answering (TKGQA)} aims to infer answers to time-sensitive questions $q \in \mathcal{Q}$ using temporal facts $f = (s, p, o, t)$ from $\mathcal{G}$. 
TKGQA requires both semantic understanding and temporal reasoning, explicitly considering temporal constraints in questions and knowledge.

\begin{figure*}[!t]
    \centering
    \includegraphics[width=0.9\textwidth]{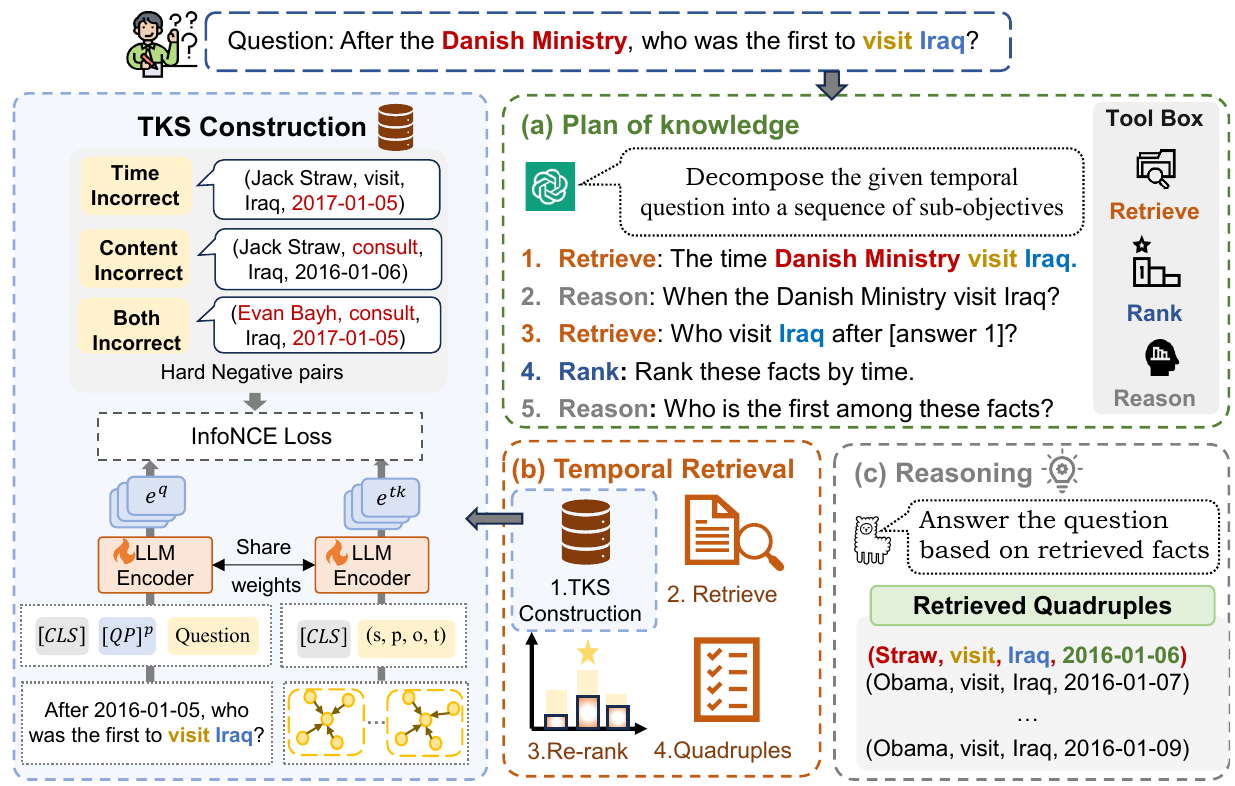}
    \caption{The overall architecture of PoK consists of three main modules: Plan of Knowledge, Temporal Retrieval, and Reasoning. The left part illustrates the construction of the Temporal Knowledge Store (TKS), which serves as the foundation for retrieval.
    }
    \label{fig: model}
\end{figure*}

\section{Method} \label{Method}
\subsection{Overview}  
We propose PoK, a time-aware Plan of Knowledge framework for complex temporal reasoning in TKGQA. 
The core idea is to plan the reasoning process by decomposing each temporal question into a sequence of sub-objectives guided by predefined operators.  
To jointly capture semantic relevance and temporal consistency, we design a temporal retrieval module using a prompt-based contrastive framework with hard negatives generated by corrupting entities, relations, or timestamps.
Task-specific prompts guide the model to focus on time-dependent reasoning constraints, and the retrieved facts are stored in a Temporal Knowledge Store (TKS) for efficient access. 
Next, a re-ranking module is used to refine candidate facts, enhancing temporal factual precision. 
Finally, the top-ranked facts are passed to the Reasoning stage, where the LLMs integrates the facts and generates the answer.

\subsection{Plan of Knowledge Framework}  
To systematically handle multi-hop and time-sensitive reasoning, we introduce a time-aware \textit{Plan of Knowledge (PoK)} framework, which explicitly models the reasoning process as a structured plan.  
The key idea is to decompose a complex temporal question into a sequence of \textit{sub-objectives} that correspond to interpretable reasoning steps.  

Formally, given a temporal question $q$, we prompt the LLM to generate a structured plan of sub-objectives as in \eqref{eq:llm}, which serves as guidance for retrieval, rank and reasoning.  
The prompt template used for this decomposition is detailed in \tbref{tab:plan}.  
\begin{equation} \label{eq:llm}
\begin{split}
O_q &= \{(op_i, o_i)\}_{i=1}^n = \text{LLM}(\text{Prompt}(q)), \\
q &\in \mathcal{Q}, \quad op_i \in \{\textit{Retrieve}, \textit{Rerank}, \textit{Reason}\}.
\end{split}
\end{equation}

where $O$ denotes the generated plan objective and each $o_i$ represents an individual sub-objective.  
Each sub-objective is grounded in a set of pre-defined temporal planning operators $op$, ensuring that the reasoning process remains structured and semantically consistent:  
\begin{itemize}
    \item \textit{Retrieval} (Section \ref{retrieve}): a mapping  
    $f_{\text{Retrieve}}: \mathcal{Q} \to 2^{\mathcal{F}}$,  
    which retrieves a subset of potentially relevant temporal facts $\mathcal{F}$ for a given sub-objective $q_i$. Here, $2^{\mathcal{F}}$ denotes the power set of $\mathcal{F}$.  

    \item \textit{Rank}: a temporal sorting function  
    $f_{\text{Rank}}: 2^{\mathcal{F}} \to \mathcal{F}^*$,  
    which orders the retrieved facts chronologically (ascending or descending), yielding an ordered fact sequence $\mathcal{F}^*$.  

    \item \textit{Reasoning} (Section \ref{reason}): a compositional inference function  
    $f_{\text{Reason}}: \mathcal{F}^* \to \mathcal{A}$,  
    which integrates the temporally ordered evidence and infers the final answer $a \in \mathcal{A}$.  
\end{itemize}  

By restricting the reasoning space to predefined temporal operators, PoK ensures that generated sub-objectives faithfully reflect the temporal semantics of the question. 
Unlike conventional question decomposition methods~\cite{rtqa} that merely split questions into sub-questions, our task-oriented planning framework explicitly grounds each sub-objective in an executable operation.

\begin{table}[htbp]
\centering
\begin{tabular}{p{0.9\linewidth}}
\hline
\textbf{Plan of Knowledge Prompt} \\
\hline
\textbf{Task}: Decompose the given temporal question into a sequence of sub-objectives. \\

\textbf{Guidelines}: \\
1. Each sub-objective must use one of the predefined operators: Retrieve, Rank, or Reason. \\
2. Separate each sub-objective with a semicolon (;). \\  
3. Use [answer i] to refer to the output of a previous sub-objective i. \\
4. Ensure the sub-objectives form a logical reasoning chain. \\

\textbf{Example}: \\
Input: Who investigated China first after Segolene Royal? \\
Output: \\
\textbf{Retrieve}: When did Segolene Royal investigate China?; \\
\textbf{Retrieve}: Who investigated China after [answer 1]?;  \\
\textbf{Rank}: Rank these facts by time; \\
\textbf{Reason}: Who is the first among [answer 2]? \\

\textbf{Now process}: \\ 
Question: \texttt{<question>} \\
\hline
\end{tabular}
\caption{Plan of Knowledge Prompt Template.}
\label{tab:plan}
\end{table}

\subsection{Temporal retrieval Module}\label{retrieve}

Existing KGQA systems~\cite{chen2023multi,liu2023local} typically follow a pipeline paradigm: entity-linking tools identify entities and relations in the question, and a retrieval module searches for candidate facts. While effective in static settings, this approach struggles on some TKGs, such as ICEWS~\cite{garcia2018learning}, where entity-linking tools are unavailable~\cite{chen2023multi}. To address this, we adopt the direct retrieval paradigm~\cite{baek2023direct}, which matches questions directly with factual embeddings.

Building on this idea, we propose a prompt-based contrastive temporal retrieval framework that captures both semantic similarity and temporal constraints. The temporal retrieval module first applies prompt-guided temporal encoding to inject task prompts into question representations, capturing temporal constraints. It then performs contrastive time-aware fine-tuning to align semantically and temporally enriched question embeddings with factual embeddings. Temporal facts are encoded into the Temporal Knowledge Store (TKS) for efficient access, followed by retrieval and re-ranking to obtain the final candidate facts.

\textbf{Prompt-guided Temporal Encoding.}
To enhance the encoder's sensitivity to temporal constraints, we integrate task soft prompts inspired by prompt learning~\cite{lester2021softprompt}. These learnable prompt tokens are inserted into the input sequence and guide the model's attention toward temporal constraints. Specifically, the prompts are represented as a sequence of learnable vectors:
$
\mathbf{P} = [\mathbf{p}_1, \mathbf{p}_2, \ldots, \mathbf{p}_m], \quad \mathbf{p}_i \in \mathbb{R}^d,
$
where $m$ denotes the number of prompt tokens. Given a question $q = [w_1, w_2, \ldots, w_n]$, the input sequence is formed as:
$
\mathbf{X} = [\mathbf{p}_q, \mathbf{e}_{w_1}, \ldots, \mathbf{e}_{w_n}],
$
and the encoded question embedding is obtained as:
\begin{equation} \label{eq:q_lm}
\mathbf{E}_{q_i} = LM_t(\mathbf{X}) \in \mathbb{R}^d.
\end{equation}

\textbf{Contrastive Time-aware Fine-tuning.}
Existing off-the-shelf retrieval tools such as BM25 rely on semantic similarity between the question and candidate facts, without considering the temporal constraints.
To address this, we introduce a contrastive learning objective that jointly optimizes semantic and temporal alignment in a shared embedding space.

For each positive question–fact pair, we construct three types of negative samples to model temporal and semantic discrepancies: time-incorrect facts by replacing timestamps, content-incorrect variants by modifying relations, and both-incorrect samples by altering entities and timestamps simultaneously. 
This strategy encourages the model to capture fine-grained temporal cues and distinguish facts that are semantically plausible but temporally invalid.
Following~\cite{son2023timeaware}, we adopt the InfoNCE loss as:
\begin{equation} \label{eq:loss}
\mathcal{L} = -\log\frac{e^{(\phi(s,p,o,t))/\tau}}{e^{(\phi(s,p,o,t))/\tau}+\sum_{o'\in\mathcal{N}^o}e^{\phi(s,p,o',t)/\tau}},
\end{equation}
where $\phi(s,p,o,t)$ denotes the cosine similarity between the question embedding and its corresponding factual embedding, $\tau$ controls the temperature of the contrastive distribution, and $\mathcal{N}^o$ is the set of negative samples.

\textbf{Temporal Knowledge Store Construction.}
To support efficient retrieval, we construct a Temporal Knowledge Store (TKS) that encodes both semantic and temporal information for facts in the TKGs. 
Specifically, each quadruple $(s, p, o, t)$ is converted into a textual template according to its temporal type. 
For time-point quadruple, the template is \texttt{"\{subject\} \{predicate\} \{object\} at \{time\_point\}"}, and for time-interval quadruple, it is \texttt{"\{subject\} \{predicate\} \{object\} from \{begin\_time\} to \{end\_time\}"}. 
After encoding, all temporal facts $T(s,p,o,t) \in \mathcal{G}$ are stored as:
\begin{equation} \label{eq:time_lm}
\text{TKS} = \{\mathbf{E}_{f} \mid \mathbf{E}_{f} = LM_t(T(s,p,o,t)), (s,p,o,t) \in \mathcal{G}\}.
\end{equation}
This dense repository acts as a neural memory that can be efficiently queried through vector similarity search. 
By representing all facts in a unified embedding space, TKS provides a continuous interface between queries and structured temporal knowledge, enabling fast and accurate retrieval during inference.

\textbf{Temporal retrieval and reranking.}  
During inference, relevant temporal facts are retrieved by computing the cosine similarity between the encoded question $\mathbf{E}_q$ and all fact embeddings $\mathbf{E}_f$ stored in TKS:
\begin{equation} \label{eq:cosine}
\phi_{\textit{TKS}} = \cos(\mathbf{E}_{q_i}, \mathbf{E}_f).
\end{equation}
For scalability, we employ FAISS~\cite{faiss} for efficient vector search and indexing. 

To improve temporal relevance, a time-filtering function is applied for questions with time constraints. For each quadruple $(s,p,o,t)$ in the TKG $\mathcal{G}$, we compute the time difference from the query time $t_q$, filter out quadruples outside the valid range, and normalize the differences to produce the filtered results. Equation~\eqref{eq:filter} illustrates the function for "before"-type questions.

\begin{equation} \label{eq:filter}
\phi_t(t_q, t) =
\begin{cases}
1 - \frac{|t_q - t|}{\max(t_q - t)}, & \text{if } (t_q - t) > 0, \\
-100, & \text{otherwise}.
\end{cases}
\end{equation}
The final retrieval score jointly integrates semantic and temporal signals:
\begin{equation} \label{eq:rerank}
\phi({q}, t) = \mu \cdot \phi_{\textit{TKS}}(\mathbf{E}_{q_i}, \mathbf{E}_t) + (1-\mu) \cdot \phi_t(t_q, t),
\end{equation}
\begin{equation} \label{eq:sim}
\mathbf{f} = \arg \max \phi({q_i}, t),
\end{equation}
where $\mu$ is a balancing coefficient controlling the trade-off between semantic relevance and temporal coherence. This integrated scoring mechanism ensures that the retrieved evidence is not only linguistically plausible but also temporally valid.

\subsection{Reasoning} \label{reason}
After retrieving the temporal facts, we cast the reasoning process as an optimization problem over the LLM, where the goal is to maximize the likelihood of producing the correct answer $a$ given the question $q$ and its retrieved evidence. Formally, the objective is defined as:
\begin{equation} \label{eq:reason}
\mathcal{L} = \max_{\Phi} 
\sum_{({q_i},a) \in \hat{\mathcal{Q}}} 
\sum_{t=1}^{|a|} 
\log P_{\Phi}\left(a_t \mid ({q_i}, f^+), a_{<t}\right),
\end{equation}
where $\Phi$ denotes the parameters of the LLM, $(q^*, f^+)$ represents the question $q$ paired with its most relevant fact, and $a_{<t}$ is the partial sequence generated up to step $t{-}1$. This formulation allows the model to jointly reason over both the context of the question and the factual knowledge retrieved from the TKS.  
To guide the LLM in generating final answers, we design a simple instruction prompt in \tbref{tab:reason}.

\begin{table}[htbp]
\centering
\begin{tabular}{p{0.9\linewidth}}
\hline
\textbf{Reasoning Prompt} \\
\hline
Based on the facts, please answer the given question. \\
Keep the answer as simple as possible and return all the possible answers as a list.\\
Facts: \texttt{<facts>} \\
Question: \texttt{<question>} \\
\hline
\end{tabular}
\caption{Reasoning Prompt Template.}
\label{tab:reason}
\end{table}

\section{Experiments} \label{Experiments}

In this section, we first describe the experimental setup, including the datasets, evaluation metrics, and baseline methods. Then, we conduct extensive experiments on two real-world datasets to answer the following research questions (RQs):

\begin{itemize}
    \item RQ1: Does PoK outperform existing baselines on temporal question answering tasks?
    \item RQ2: How does PoK generalize across different LLMs?
    \item RQ3: What is the contribution of each component to the overall performance ofPoK?
    \item RQ4: How does PoK perform in complex questions?
    \item RQ5: How effective is the temporal retrieval in PoK compared with other retrieval strategies?
    \item RQ6: How do the number of facts impact the performance of PoK?
    \item RQ7: How does the efficiency of PoK compare to state-of-the-art baselines?
\end{itemize}

\subsection{Experiment Settings}
\subsubsection{Datasets}
To evaluate the performance of PoK, we conduct experiments on four popular datasets, 
 MULTITQ~\cite{chen2023multi}, TimeQuestions~\cite{jia2021complex}, Timeline-ICEWS, and Timeline-CronQuestions~\cite{timeline}.
 The statistical information is presented in \tbref{tab:data_stats}.

\textbf{MULTITQ} is the largest publicly available TKGQA dataset, constructed from ICEWS05-15~\cite{icews}, and contains more than 500K unique question–answer pairs. 
A distinctive feature of MULTITQ is that it covers multiple temporal granularities, including year, month, and day, with time spans ranging from 2005-01-01 to 2015-12-31.

\textbf{TimeQuestions} is another widely used and challenging TKGQA benchmark, constructed from temporal facts in Wikidata. Each fact is represented in the form of \{subject, predicate, object, begin\_time, end\_time\}, thereby providing rich temporal information for reasoning.
Compared with MULTITQ, TimeQuestions is smaller in size and only supports year-level temporal granularity. Nevertheless, it spans a much broader temporal range, covering facts across more than 1,600 years.

\textbf{Timeline-ICEWS} and \textbf{Timeline-CronQuestion} are 
datasets constructed from ICEWS Coded Event Data (Time Range) and CronQuestion knowledge graph (Time Point) using TimelineKGQA categorization framework. 
% They contain complex questions through multiple dimensions: (1) context complexity (Simple, Medium, Complex), (2) answer focus (Temporal vs. Factual), (3) temporal relations (Thirteen Allen Temporal Relations, Three Time Range Set Relations, Duration Comparisons and Ranking), and (4) temporal capabilities, grouped into temporal retrieval capabilities (temporal constrained retrieval and timeline position retrieval) and temporal operation capabilities (temporal semantic operation and timeline arithmetic operation). 
\begin{table}[ht]
\centering
\begin{tabular}{l l r r r}
\toprule
\textbf{Dataset} & \textbf{Type} & \textbf{Train} & \textbf{Val} & \textbf{Test} \\
\midrule
\multirow{2}{*}{MULTITQ} 
& Single & 283,482 & 41,735 & 38,864 \\
& Multiple & 103,305 & 16,244 & 15,720 \\
\cmidrule{2-5}
& Total & 386,787 & 57,979 & 54,584 \\
\midrule

\multirow{4}{*}{TimeQuestions} 
& Explicit & 3,904 & 1,302 & 1,311 \\
& Implicit & 871 & 299 & 292 \\
& Temporal & 3,222 & 1,073 & 1,067 \\
& Ordinal & 1,711 & 570 & 567 \\
\cmidrule{2-5}
& Total & 9,708 & 3,236 & 3,237 \\
\midrule

\multirow{3}{*}{Timeline-ICEWS} 
& Simple & 17,982 & 5,994 & 5,994 \\
& Medium & 15,990 & 5,330 & 5,330 \\
& Complex & 19,652 & 6,550 & 6,550 \\
\cmidrule{2-5}
& Total & 53,624 & 17,874 & 17,874 \\
\midrule

\multirow{3}{*}{Timeline-CronQuestion} 
& Simple & 7,200 & 2,400 & 2,400 \\
& Medium & 8,252 & 2,751 & 2,751 \\
& Complex & 9,580 & 3,193 & 3,193 \\
\cmidrule{2-5}
& Total & 25,032 & 8,344 & 8,344 \\
\bottomrule
\end{tabular}
\caption{Data splits for the datasets.}
\label{tab:data_stats}
\end{table}

\subsubsection{Parameter Setting}

For the \textbf{planning framework}, we leverage the OpenAI API\footnote{\href{https://platform.openai.com/docs/api-reference}{https://platform.openai.com/docs/api-reference}}
 with the gpt-4o\footnote{\href{https://platform.openai.com/docs/models/gpt-4o}{https://platform.openai.com/docs/models/gpt-4o}}
 model to decompose questions into sub-objectives that guide temporal reasoning.
For the \textbf{temporal retriever}, we fine-tune Qwen3-Embedding-0.7B~\cite{qwen3} for 2 epochs, which offers a lightweight yet highly effective solution for semantic retrieval. Each training question is optimized with in-batch negatives as well as three hard negatives.
We set the temperature $\tau=0.01$. 
During the re-ranking stage, we introduce a balancing coefficient $\mu=0.2$ to combine semantic relevance with temporal consistency.
For the \textbf{reasoning backbone}, we conduct a comparative study among several LLMs, including GPT-4o, Qwen3-8B, and LLaMA2-Chat-7B~\cite{touvron2023llama}. Based on both performance and efficiency considerations, we ultimately adopt LLaMA2-Chat-7B as our reasoning backbone. The model is fine-tuned for 2 epochs on 2 NVIDIA A6000 GPUs. Due to the considerable size of the MULTITQ dataset, we sample only 20\% of its training set for fine-tuning. During inference, we further enhance reasoning by retrieving the top-20 candidate facts from the retriever and feeding them into the LLM.
Evaluation metrics are Hits@k. Hits@k refers to the percentage of correct relations ranked in the top k predictions. Higher Hits@k demonstrate better performance.

\subsubsection{Baselines}
To validate the effectiveness of the proposed model, we compare the proposed PoK with three groups of baselines  on MULTITQ:

\begin{itemize}
    \item \textbf{Pre-trained LM-based methods}: This category includes BERT~\cite{devlin2019bert}, DistilBERT, ALBERT~\cite{lan2020albert}. For BERT and its variants, we generate LM-based question embeddings, concatenate them with entity and temporal embeddings, and apply a learnable projection layer, following previous work~\cite{chen2023multi}.
    
    \item \textbf{TKGQA methods}: This category includes EmbedKGQA~\cite{improving}, CronKGQA~\cite{saxena2021question}, and MultiQA~\cite{chen2023multi}. 
    % EmbedKGQA is designed for static KGs, CronKGQA focuses on single-granularity temporal reasoning, and MultiQA leverages transformers to model temporal information across multiple granularities. All these methods rely on temporal KG embeddings and use scoring functions to identify appropriate answers.
    
    \item \textbf{LLM-based methods}: This category includes LLaMA2-7B, GPT-4o, ARI~\cite{chen2023temporal}, Naive RAG~\cite{chen2023benchmarkinglargelanguagemodels}, ReAct RAG Agent~\cite{yao2023reactsynergizingreasoningacting}, TempAgent~\cite{timeagent}, Time$R^4$~\cite{timer4}, and RTQA~\cite{rtqa}.
     For LLaMA2 and GPT-4o, we directly feed the questions as input without additional explanations.
    % Naive RAG represents one of the earliest retrieval-augmented generation approaches after the emergence of GPT-4o. ReAct provides a simple yet effective framework for interpretable, step-by-step reasoning and decision-making in LLMs. TempAgent introduces an autonomous agent paradigm that enhances temporal reasoning capabilities. Time$R^4$ reformulates questions to improve temporal understanding, while RTQA decomposes questions into sub-problems for more effective reasoning.

\end{itemize}

We also select three groups of baselines for comparison on TimeQuestions:

\begin{itemize}
    \item \textbf{Static KGQA methods}: including PullNet~\cite{sun2019pullnet}, Uniqorn~\cite{pramanik2023uniqorn}, and GRAFT-Net~\cite{sun2018open}. 
    % GRAFT-Net was the first technique to adapt R-GCNs for QA over heterogeneous sources. 
    % PullNet extended the GRAFT-Net classifier to the scenario of multi-hop questions. 
    % Uniqorn is a method for answering complex questions using Group Steiner Trees.
    
    \item \textbf{TKGQA methods}: including CronKGQA~\cite{saxena2021question}, TempoQR~\cite{mavromatis2022tempoqr}, EXAQT~\cite{jia2021complex}, LGQA~\cite{liu2023local}, TwiRGCN~\cite{sharma2022twirgcn}, TMA~\cite{liu2023timeawaremultiwayadaptivefusion}, and $M^3$TQA~\cite{m3tqa}.
    % TempoQR enhances question embeddings by integrating contextual, entity-level, and time-aware features through three dedicated modules.  
    % TwiRGCN~\cite{sharma2022twirgcn} introduces temporally weighted graph convolutions followed by answer gating to emphasize relevant temporal edges. 
    % LGQA~\cite{liu2023local} employs a multi-hop message passing Graph Neural Network to integrate both local and global temporal information. 
    % $M^3$TQA is a novel Multi-view, Multi-hop and Multi-stage reasoning paradigm.
    
    \item \textbf{LLM-based methods}: This category includes LLaMA2-7B, GPT-4o, GenTKGQA\cite{gentkgqa}, and Time$R^4$~\cite{timer4}.
    % GenTKGQA employs a pre-trained T-GNN layer to embed this subgraph into virtual indicators for question representation.
    For LLaMA2 and GPT-4o, we also directly feed the questions as input without additional explanations.

\end{itemize}

For timeline-ICEWS and timeline-CronQuestions, we compare our model with the latest RTQA and RAG baselines. The RAG baseline is implemented by encoding both the knowledge and queries using OpenAI's text-embedding-3-small model, followed by a semantic similarity search to retrieve the top-1 most relevant fact.
 
\begin{table}[t]
\centering
\small
\resizebox{\columnwidth}{!}{%
\begin{tabular}{lcccccc}
\toprule
Model & Overall & \multicolumn{2}{c}{Question Type} & \multicolumn{2}{c}{Answer Type} \\
\cmidrule(lr){3-4} \cmidrule(lr){5-6}
& & Multiple & Single & Entity & Time \\
\midrule
BERT & 8.3 & 6.1 & 9.2 & 10.1 & 4.0 \\
DistillBERT & 8.3 & 7.4 & 8.7 & 10.2 & 3.7 \\
ALBERT & 10.8 & 8.6 & 11.6 & 13.9 & 3.2 \\
\midrule
EmbedKGQA & 20.6 & 13.4 & 23.5 & 29.0 & 0.1 \\
CronKGQA & 27.9 & 13.4 & 33.7 & 32.8 & 15.6 \\
MultiQA & 29.3 & 15.9 & 34.7 & 34.9 & 15.7 \\
\midrule
LLaMA2-7B & 18.5 & 10.1 & 22.0 & 23.9 & 5.5 \\
ChatGPT & 10.2 & 7.7 & 14.7 & 13.7 & 2.0 \\
ARI & 38.0 & 21.0 & 68.0 & 39.4 & 34.4 \\
Naive RAG & 37.9 & 15.5 & 46.9 & 24.2 & 67.2 \\
ReAct RAG & 39.8 & 13.0 & 50.6 & 24.3 & 73.5 \\
TempAgent & 70.2 & 31.6 & 85.7 & 62.4 & 87.0 \\
Time$R^4$ & 72.8 & 33.5 & 88.7 & 63.9 & 94.5 \\
RTQA & \underline{76.5} & \textbf{42.4} & \underline{90.2} & \underline{69.2} & \underline{94.2} \\
\midrule
PoK & \textbf{77.9} & \underline{40.9} & \textbf{92.9} & \textbf{69.6} & \textbf{96.2} \\
\bottomrule
\end{tabular}%
}
\caption{Hits@1 performance comparison of different models on MULTITQ (\%).}
\label{tab:multi}
\end{table}

\begin{table}[t]
\centering
\small
\resizebox{\columnwidth}{!}{%
\begin{tabular}{lcccccc}
\toprule
\multirow{2}{*}{Model} 
& \multirow{2}{*}{Overall} & \multicolumn{2}{c}{Question Type} & \multicolumn{2}{c}{Answer Type}  \\
\cline{3-4} \cline{5-6} 
& & Explicit & Implicit & Temporal & Ordinal \\
\hline
PullNet  & 10.5  & 2.2  & 8.1  & 23.4  & 2.9 \\
Uniqorn  & 33.1  & 31.8  & 31.6  & 39.2  & 20.2 \\
GRAFT-Net  & 45.2  & 44.5  & 42.8  & 51.5  & 32.2 \\
\hline
CronKGQA  & 46.2  & 46.6  & 44.5  & 51.1  & 36.9 \\
TempoQR  & 41.6  & 46.5  & 3.6  & 40.0  & 34.9 \\
EXAQT  & 57.2  & 56.8  & 51.2  & 64.2  & 42.0 \\
TwiRGCN  & 60.5  & 60.2  & 58.6  & 64.1  & 51.8 \\
LGQA  & 52.9  & 53.2  & 50.6  & 60.5  & 40.2 \\
TMA &  43.5 &  44.2 &  41.9 &  47.6 &  35.2 \\
$M^3$TQA &  53.6 &  53.6 &  51.0 &  61.2 &  40.8 \\
\hline
LLaMA2-7B  & 27.1 & 26.8 & 32.5 & 27.9 & 23.4 \\
ChatGPT    & 45.9 & 43.3 & 51.1 & 46.5 & 48.1 \\
GenTKGQA &  58.4  & 59.6 &  61.1 &  56.3 &  57.8 \\
TimeR$^{4}$ & \underline{78.1} & \underline{82.3} & \underline{73.0} & \underline{83.0} & \underline{64.9} \\
\hline
PoK & \textbf{83.2} & \textbf{85.8} & \textbf{85.3} & \textbf{74.3} & \textbf{84.2}  \\
\hline
\end{tabular}
}
\caption{Hits@1 performance comparison of different
models on TimeQuestions (\%).}
\label{tab:timequestion}
\end{table}

\begin{table*}[t]
\centering
\small
\begin{tabular}{lccccccccc}
\toprule
\multirow{2}{*}{Model} & \multicolumn{4}{c}{CronQuestions KG} & \multicolumn{4}{c}{ICEWS Actor} \\
\cmidrule(lr){2-5} \cmidrule(lr){6-9}
& Overall & Simple & Medium & Complex  & Overall & Simple & Medium & Complex  \\
\midrule
RAG baseline  &  23.5  &  70.4  &  9.2  &  0.9   &  26.5   &  66.0   &  12.8   &  1.1 \\
LLaMA2-7B & 16.9  & 4.9  & 14.3  & 28.2   & 11.1  & 3.5  & 6.6  & 32.2 \\
GPT-4o & 20.6 & 6.9  & 13.0  & 37.6  & 11.3 & 5.1 & 3.5 & 35.3  \\

RTQA  &   \underline{29.8}  &   \underline{60.8}  &   \underline{21.8}  &   \underline{13.5}  & - & - & - & - \\
\midrule
PoK  & \textbf{65.1} & \textbf{73.7} & \textbf{53.9} & \textbf{68.3} & \textbf{60.2} & \underline{74.4}  & \textbf{45.6} & \textbf{57.8} \\
\bottomrule
\end{tabular}
\caption{Performance comparison of different models (in percentage) on TimelineKGQA. }
\label{tab:timeline}
\end{table*}
\begin{table*}[!t]
\centering
\resizebox{1\linewidth}{!}{
\setlength{\tabcolsep}{1.5mm}{
\begin{tabular}{lcccccccccc}
\hline
\multirow{2}{*}{Model} & \multicolumn{5}{c}{MULTITQ} & \multicolumn{5}{c}{Timequestions} \\
\cline{2-6} \cline{7-11}
& Overall & Single & Multiple & Entity & Time & Overall & Explicit & Implicit & Ordinal & Temporal \\
\hline
PoK  & \textbf{77.9} & \textbf{92.9} & \textbf{40.9} & \textbf{69.6} & \textbf{96.2} & \textbf{83.2} & \textbf{85.8} & \textbf{85.3} & \textbf{74.3} & \textbf{84.2} \\
\hline
w/o PoK-plan & \underline{71.3} & \underline{91.1} & 22.2 & \underline{62.6} & 92.5 & \underline{82.4} & \underline{85.7} & \underline{79.1} & \underline{73.7} & \underline{84.0} \\
w/o PoK-rank  & 75.5 & 92.3 & 34.1 & 67.3 & 95.5  & 81.1  & 84.4  & 70.6  & 71.8  & 83.7 \\
w/o PoK-retrieve  & 32.0  & 36.3  & 21.2  & 39.7  & 13.1  & 44.7  & 44.4  & 55.5  & 45.0  & 42.0 \\
w/o task prompt & 74.9 & 90.0 & 37.8 & 68.3 & 91.1 & 82.1  & 85.0  & 77.4  & 74.1  & 83.4 \\
w/o re-rank & 71.0 & 87.4 & \underline{30.5} & 62.2 & \underline{92.6} & 79.7 & 84.4 & 57.3 & 72.8 & 83.0 \\
\hline
\end{tabular}
}}
\caption{Results of the ablation study. “w/o” means removing the module.}
\label{tab:ablation}
\end{table*}

\begin{table*}[!t]
\centering
\resizebox{\linewidth}{!}{
\setlength{\tabcolsep}{1.5mm}{
\begin{tabular}{lcccccccccc}
\hline
\multirow{2}{*}{Model} & \multicolumn{5}{c}{MULTITQ} & \multicolumn{5}{c}{Timequestions} \\
\cline{2-6} \cline{7-11}
& Overall & Single & Multiple & Entity & Time & Overall & Explicit & Implicit & Ordinal & Temporal \\
\hline
PoK  &\textbf{77.9} & \textbf{40.9} & \textbf{92.9} & \textbf{69.6} & \textbf{96.2}  & \textbf{83.2} & \textbf{85.8} & \textbf{85.3} & \textbf{74.3} & \textbf{84.2}   \\

LLaMA2-7B  & 18.5 & 22.0  & 10.1 &  23.9  & 5.5 & 28.9 & 26.8 & 41.9 & 33.7 & 33.8 \\
LLaMA2 \textit{w/ finetuned} & 33.9 & 38.4 & 22.7 & 45.0 & 7.8  &  45.8 & 44.4 & 46.0 & 51.9 & 37.8\\
GPT-4o       & 10.2 & 14.7 & 7.7 & 13.7 & 2.0  & 45.9 & 43.3 & 51.1 & 46.5 & 42.1 \\

\hline
% LLaMA2 \textit{w/ lora}  & - & - & - & - & - & 74.3 & 76.9 & 72.6 & 81.8 & 54.5 \\

LLaMA2-7B \textit{w/ PoK }  &  58.4 & 39.5   & 65.6  & 60.9  & 52.3  &  61.3  & 62.5  & 53.4  & 36.2  & 75.4 \\
Qwen3-8B \textit{w/  PoK}    & 63.1  & 36.6  & 73.9  &  66.4  & 55.2 & 64.8  & 67.5  & 53.1  & 39.5  & 77.3 \\
GPT-4o \textit{w/  PoK}    & \underline{66.0}  & \underline{39.6}  & \underline{76.7}  & \underline{67.5}  & \underline{62.3} & \underline{67.1}  & \underline{70.2}  & \underline{57.5}  & \underline{45.5}  & \underline{77.3} \\

\hline
\end{tabular}
}}
\caption{Effects of integrating the PoK framework with different LLMs for reasoning.}
\label{tab: LLMs}
\end{table*}

\subsection{Main Results (RQ1)}
We present the experimental results in comparison with other methods on the MULTITQ, TimeQuestions, timeline-CronQuestions, and timeline-ICEWS datasets, as shown in \tbref{tab:multi}, \tbref{tab:timequestion}, and \tbref{tab:timeline}. 
In these tables, the best results are highlighted in bold, while the second-best results are underlined. 
Across all experimental settings, PoK consistently achieves the best performance, clearly demonstrating its effectiveness and robustness in addressing the TKGQA task.

For the MULTITQ dataset, PoK achieves state-of-the-art performance. 
Specifically, we find that PLMs (BERT. ALBERT) and LLMs (LLaMA2, GPT-4o) exhibit the lowest performance on the TKGQA task. This might be due to the lack of necessary temporal knowledge, thus leading to errors in reasoning.
Compared to traditional KGQA methods, PoK achieves a 62.3\% relative improvement in Hits@1, underscoring its capability to handle temporally complex queries. 
Moreover, when compared with recent LLM-based methods such as ARI, RTQA, and TempAgent, PoK achieves relative improvements of 51.2\% and 1.8\%, respectively. 
These results validate the strength of our proposed PoK framework in temporal retrieval and reasoning. 
In addition, PoK also surpasses retrieval-augmented methods such as Naive RAG and ReAct RAG, which further highlights the effectiveness of our temporal retriever.

On the TimeQuestions dataset, PoK achieves the best results across all question categories. KGQA-based methods perform the worst, as they lack the ability to retrieve or reason over temporal facts. 
LLMs such as GPT-4o and LLaMA2 perform better on TimeQuestions than on MULTITQ, likely because the dataset is constructed on Wikidata~\cite{wiki}, which is heavily represented in their pre-training corpora and provides partial knowledge. 
Moreover, PoK surpasses GenTKGQA, further demonstrating the reliability and generalizability of its temporal retriever and Plan of Knowledge framework.

On the timeline-ICEWS and timeline-CronQuestions datasets, PoK also achieves the best results across all question categories, with relative gains of 54.2\% and 56.0\%, respectively. Notably, for complex questions, our method shows substantial improvements of 80.2\% and 38.9\%, further demonstrating the superiority and effectiveness of our approach.

\subsection{Ablation Study (RQ2)}
In this section, we conducted a series of ablation studies to assess the effectiveness of the proposed model. The ablation results are shown in \tbref{tab:ablation}.

\textbf{Effect of the PoK-plan Operator.}
To examine the contribution of our reasoning component, we remove the PoK-Plan and instead directly retrieve facts using the original questions. The results show that without the framework, overall performance on Hits@1 drops by 9.3\%, clearly demonstrating the importance of our reasoning mechanism in guiding the retrieval and integration of temporal information.
The drop on TimeQuestions is less significant, mainly because:
(1) Its events are sparse, allowing direct retrieval to find relevant facts without complex reasoning; and
(2) Many LLMs are pre-trained on Wikidata, the source of TimeQuestions, so they already encode much of its factual knowledge.

\textbf{Effect of the PoK-rank Operator.} 
We further analyze the rank operator in PoK's reasoning module. Specifically, during inference, we randomly shuffle the facts instead of preserving their chronological order. The results demonstrate a clear performance degradation after shuffling, particularly for ordinal questions, indicating that maintaining the temporal order of facts is essential for enabling LLMs to effectively comprehend and utilize temporal knowledge.

\textbf{Effect of the PoK-retrieve Operator.}
To further evaluate the role of the PoK-retrieve Operator, we remove the retrieval component and let the model rely solely on the original questions without accessing temporally relevant facts. The results show a substantial drop in performance, confirming that temporal retrieval plays a critical role in supporting reasoning. By explicitly retrieving relevant temporal facts, the strategy helps LLMs ground their answers, thereby mitigating hallucinations and improving accuracy on implicit temporal questions. 

\textbf{Effect of the task prompt module.}
To evaluate the effect of task prompt module, we remove these task prompt and provide the questions to the model. The results show a clear degradation in performance, particularly on questions requiring fine-grained temporal reasoning, indicating that task prompting helps LLMs better focus on temporal constraints and enhances their ability to reason over temporal events.

\textbf{Effect of re-rank strategy.}
Removing the rerank strategy resulted in a significant decrease in model performance, indicating that filtering out irrelevant time information is indeed crucial.

\begin{figure*}[!t]
    \centering
    \includegraphics[width=0.8\textwidth]{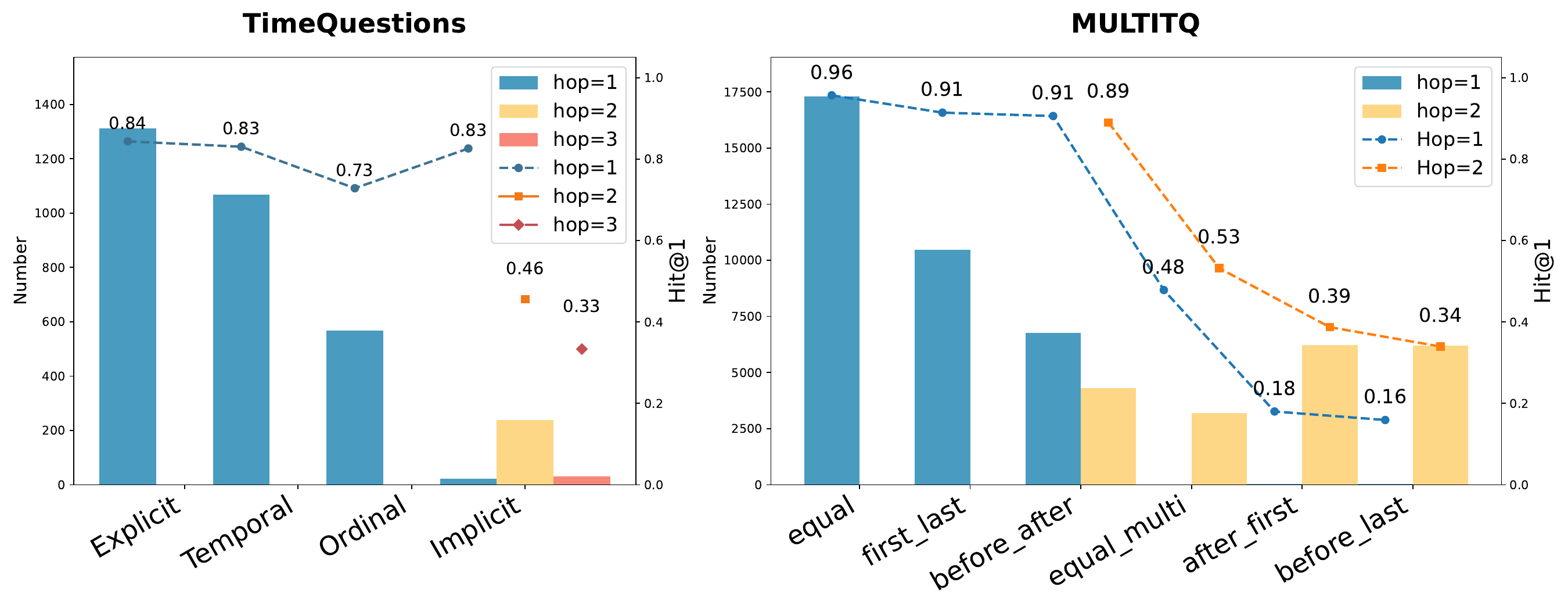}
    \caption{Comparison of Question Hops and Hit@1 Results across Different Question Types under the PoK Framework.
    }
    \label{fig: hop}
\end{figure*}
\begin{figure*}[!t]
    \centering
    \includegraphics[width=0.8\textwidth]{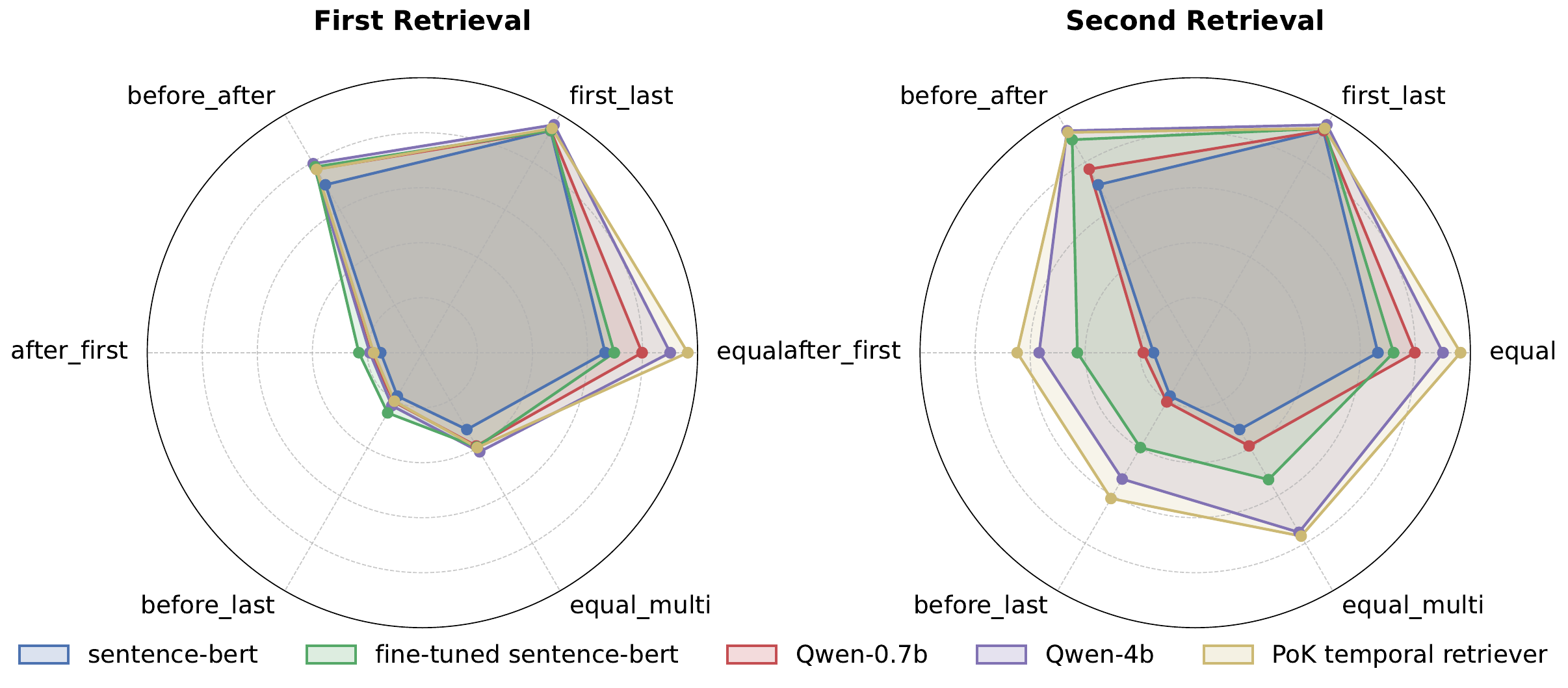}
    \caption{Comparison of Retriever Performance on the MULTITQ Dataset, including Sentence-BERT, Fine-tuned Sentence-BERT, Qwen-0.7B, Qwen-4B, and PoK Temporal Retriever.
    }
    \label{fig: rader}
\end{figure*}

\subsection{Generalizability across LLMs (RQ3)}
We compare PoK with other LLMs (LLaMA2-7B, GPT, and Qwen3) on both datasets, as shown in \tbref{tab: LLMs}. Here, LLMs w/ PoK denote backbone models equipped with PoK strategy and the planned questions.

When using LLMs alone, their performance is much higher on TimeQuestions than on MULTITQ. This is because TimeQuestions largely draws from Wikidata—part of many LLMs' pretraining corpora—allowing them to answer directly from prior knowledge. In contrast, MULTITQ involves more specialized political and event-centric temporal knowledge rarely covered in pretraining, making it more challenging.

With our PoK strategy, LLMs w/ PoK achieve substantially better results, showing that while LLMs have some inherent temporal reasoning ability, their precision and robustness are limited. PoK explicitly structures retrieval and reasoning, overcoming these limitations and consistently improving performance across backbones, demonstrating its generality.
The relatively poor performance of \textit{GPT-4o w/ PoK} possibly due to a mismatch between the output data format and the ground truth. We discuss this issue in Section~\ref{sec:case}.

We also observe that fine-tuned LLaMA2 nearly doubles the performance of the untuned model, indicating that fine-tuning effectively constrains the output space and strengthens temporal reasoning.

\subsection{Multi-hop Question Analysis (RQ4)}
We analyze the number of reasoning hops and corresponding Hit@1 results for different question types on MULTITQ and TimeQuestions, as illustrated in Figure~\ref{fig: hop}. 
We did not include an additional analysis for timeline-ICEWS and timeline-CronQuestions, since these datasets inherently categorize questions based on hop count, where simple, medium, and complex correspond to 1-hop, 2-hop, and 3-hop questions, respectively. In comparison, MULTITQ and TimeQuestions lack explicit annotations of question complexity.

The plots show that multi-hop reasoning is only required for more challenging questions—such as \textit{Implicit} in TimeQuestions and \textit{equal\_multi}, \textit{before\_last}, \textit{after\_first} in MULTITQ—while simple questions are typically handled in a single hop. This indicates that our plan of knowledge framework effectively identifies when multi-step reasoning is necessary.

From the Hit@1 trends, performance generally declines as the number of hops increases. Single-hop questions achieve consistently high accuracy, while two- and three-hop questions show a clear drop, reflecting the greater difficulty of multi-hop temporal reasoning. Nonetheless, the model still performs reasonably well on these harder cases, indicating that PoK's decomposition strategy effectively guides LLMs through complex reasoning chains.

\subsection{Effectiveness of the Retrieval (RQ5)}

To thoroughly evaluate the effectiveness of our temporal retrieval approach, we conduct a comparative analysis across several representative models, including Sentence-BERT, fine-tuned Sentence-BERT, Qwen-0.7B, Qwen-4B, and our proposed PoK temporal retriever.
Here, the fine-tuned Sentence-BERT is trained using the same data, strategies, and objectives as PoK temporal retrival to ensure a fair comparison.
We assess their answer coverage on the MULTITQ dataset and visualize the results for different categories of temporal questions using radar charts, as illustrated in \figref{fig: rader}.

The categories \textit{first\_last}, \textit{before\_after}, and \textit{equal} represent relatively simple temporal reasoning problems.
In these cases, the performance gap among different models is relatively small; however, our method still surpasses other retrievers and achieves strong results even in the first retrieval step.

In contrast, the situation changes in the second retrieval stage, which involves more complex and challenging categories such as \textit{after\_first}, \textit{before\_last}, and \textit{equal\_multi}.
Here, the overall performance improves substantially compared with the first retrieval, but more importantly, the differences between models become much more pronounced.
Notably, our fine-tuned PoK temporal retriever enables the Qwen-0.7B model to outperform the much larger Qwen-4B model, underscoring the critical role of contrastive, temporally-aware fine-tuning.

In contrast, models without fine-tuning (e.g., Sentence-BERT and Qwen-0.7B) exhibit almost no improvement across retrieval rounds.
This observation suggests that, even under the Plan of Knowledge (PoK) framework, smaller embedding models lack the capacity to effectively capture and reason over fine-grained temporal constraints (e.g., before, first).
By integrating contrastive fine-tuning, our approach explicitly encourages the embedding model to learn temporal distinctions and attend more carefully to time-sensitive relational cues.
These results collectively demonstrate that temporal-aware fine-tuning is essential for enabling retrieval models to achieve robust and accurate temporal reasoning.

\subsection{Number of Retrieved Facts (RQ6)}
\begin{figure*}[t]
    \centering
    \includegraphics[width=0.24\textwidth]{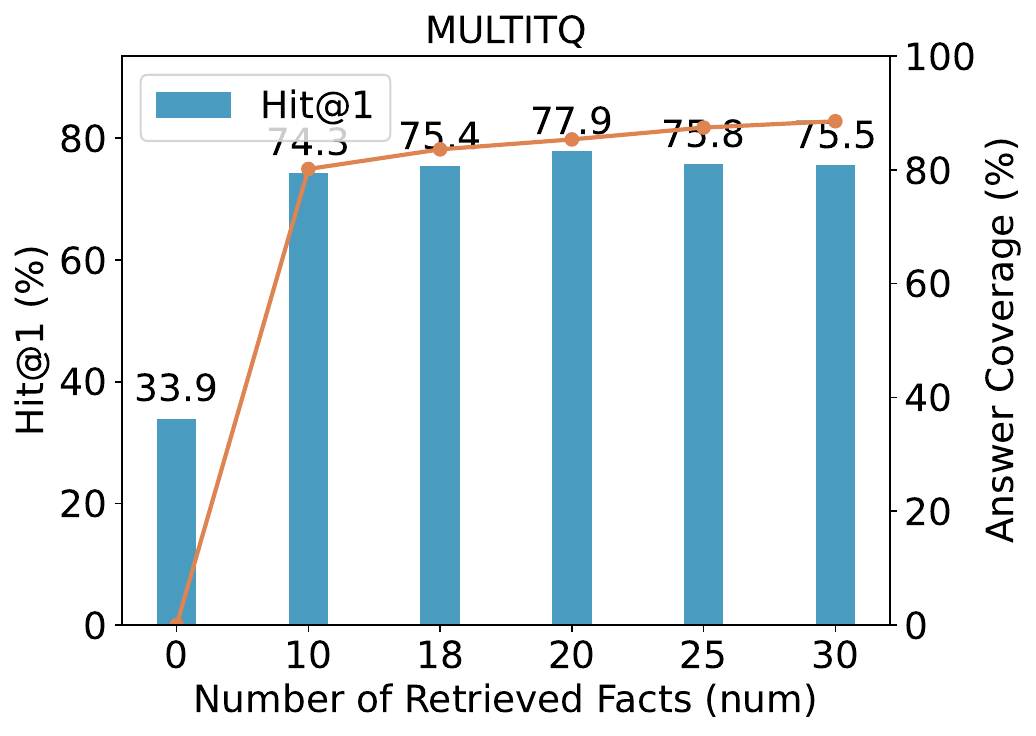}%
    \hfill
    \includegraphics[width=0.24\textwidth]{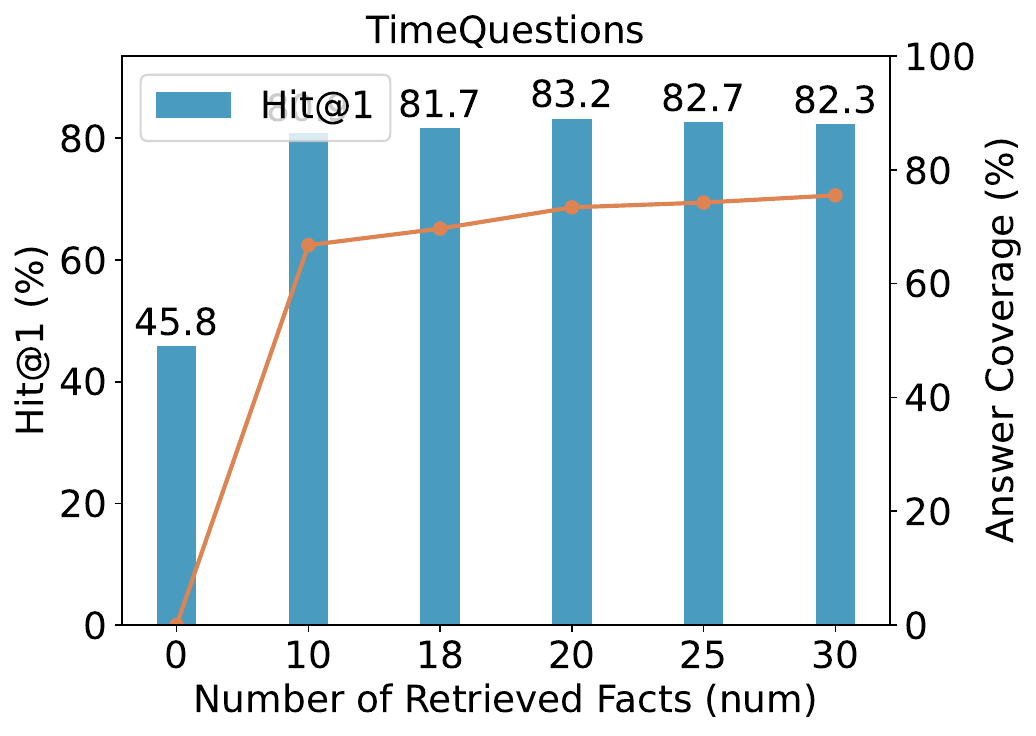}%
    \hfill
    \includegraphics[width=0.24\textwidth]{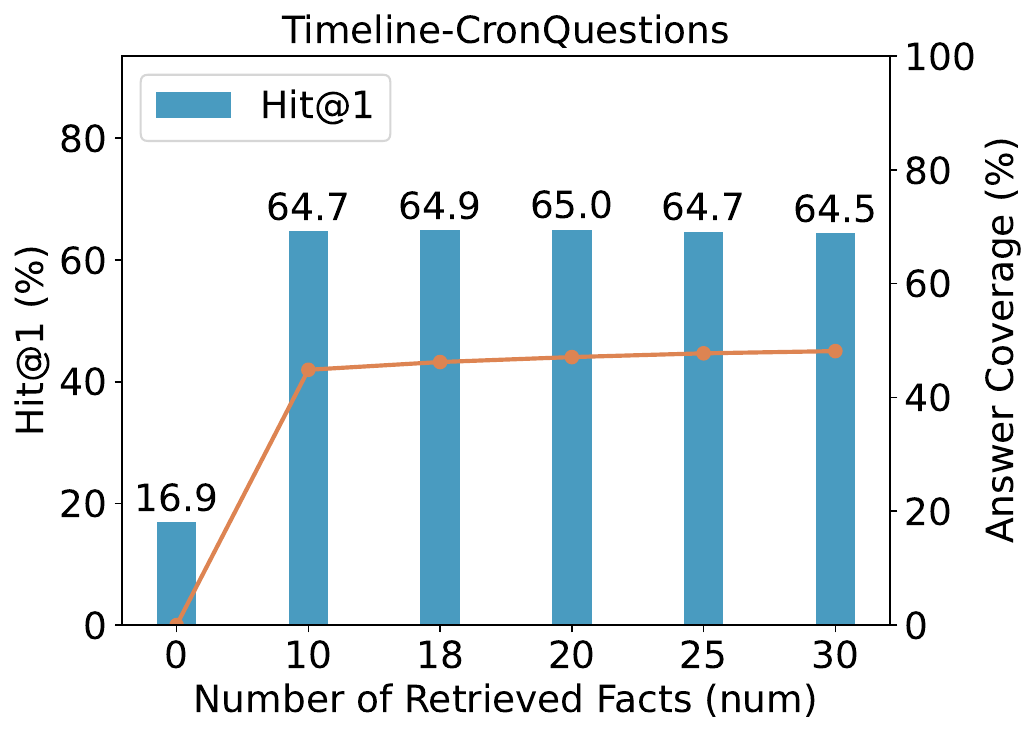}%
    \hfill
    \includegraphics[width=0.24\textwidth]{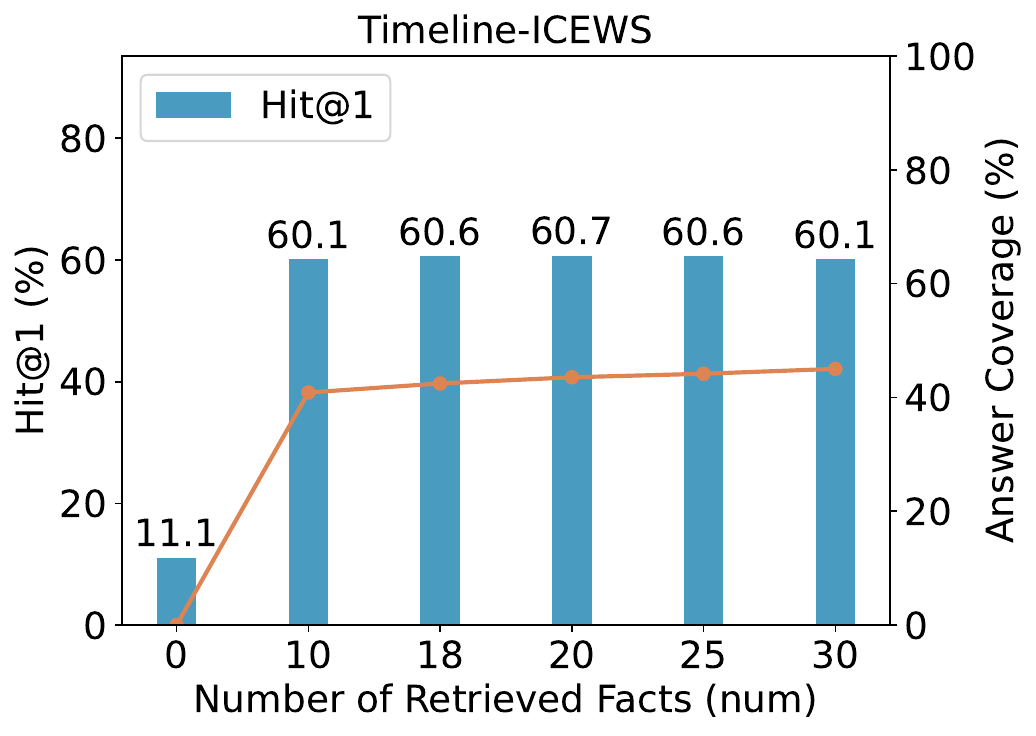}%
    \caption{
        The Hits@1 results of different retrieved fact numbers across four datasets.
    }
    \label{fig: fact}
\end{figure*}

To examine how the number of retrieved facts affects performance, we vary $n$ and report performance and answer coverage in \figref{fig: fact}.
It is evident that the model achieves its peak performance when provided with 20 relevant facts, which is the same number adopted in our retrieval strategy.
Notably, performance drops slightly at $n=25$ despite higher coverage, implying that excessive facts introduce noise, whereas too few facts limit context. Thus, $n=20$ strikes the best balance.

\begin{table}[t]
\centering
\renewcommand{\arraystretch}{1.3}
\begin{tabularx}{\columnwidth}{l|X}
\toprule
\textbf{Type} & Overlap, 3-hops \\ 
\midrule
\textbf{Question} & Who held a position in the 4th United States Congress and was Secretary of State during Andrew Jackson's presidency? \\ 
\midrule
\textbf{Answer} & Martin Van Buren, Edward Livingston, Louis McLane. \\ 
\midrule
\textbf{Plan} &
(1) Retrieve \& Reason: [time] = When is Andrew Jackson's presidency? \newline
(2) Retrieve \& Reason: [Person] = Who was the Secretary of State during [time]? \newline
(3) Retrieve \& Reason: Who held a position in the 4th United States Congress among [person]? \\
\midrule
\textbf{PoK Response} &
(1) [1829,1837] \newline
(2) ['Martin Van Buren','Edward Livingston', 
'Louis McLane','John Forsyth'] \newline
(3) ['\textbf{Martin Van Buren}', '\textbf{Edward Livingston}', '\textbf{Louis McLane}'] \\
\midrule
\textbf{GPT-4o Response} &- John Forsyth - \textbf{Martin Van Buren} \\

\bottomrule
\end{tabularx}
\caption{Example of a three-hop temporal reasoning process in PoK and GPT-4o.}
\label{tab:3hop}
\end{table}

\begin{table}[!t]
\centering

\renewcommand{\arraystretch}{1.3}
\begin{tabularx}{\linewidth}{l|X}
\toprule

\textbf{Type} & After \& first, 2-hops \\ 
\midrule
\textbf{Question} & After Okada Katsuya, who wish to visit Cambodia first? \\ 
\midrule
\textbf{Answer} & Foreign Affairs (South Korea) \\ 
\midrule
\textbf{Plan} &
(1) Retrieve \& Reason: [time] = When Okada Katsuya visits Cambodia?  \newline
(2) Retrieve: Who wishes to visit Cambodia first after [time]?  \newline
(3) Rank by timestamps in ascending order.  \newline
(4) Reason: Who wishes to visit Cambodia first after [time]? \\
\midrule
\textbf{PoK Response} &
(1) ['2009-10-02'] \newline
(2) ['Foreign Affairs (South Korea)'] \\
\midrule
\textbf{GPT-4o Response} & 
- South Korea - Thailand  -\textbf{Foreign Affairs (South Korea)} \\

\bottomrule
\end{tabularx}
\caption{Example of a two-hop temporal reasoning process in PoK and GPT-4o.}
\label{tab:2hop}
\end{table}

\subsection{Runtime Efficiency (RQ7)}
\begin{table}[t]
\centering
\resizebox{\columnwidth}{!}{%
\begin{tabular}{lccc}
\hline
Method   & Avg Time (s) & API Calls & Prompt Length\\
\hline
RTQA  &     3.58     & 3.96   &  1185 + 13.01 + 50 * 67.2 \\
PoK    &   2.28   & 1 & 134 + 13.01 + 20 * 67.2 \\
\hline
\end{tabular}
}
\caption{Comparison of API Call Counts, Runtime, and Prompt Length Across Methods.}
\label{tab:api_runtime_comparison}
\end{table}
To evaluate efficiency, we compare PoK with the latest baseline RTQA on MULTITQ in terms of average runtime, API calls, and prompt length (\tbref{tab:api_runtime_comparison}). PoK's total runtime consists of training (0.65 s), retrieval (0.06 s), planning (0.95 s), and reasoning (0.52 s). By fine-tuning only a subset of the training data, PoK not only reduces computational cost but also achieves efficient training. Moreover, inference after fine-tuning is faster than that of the base model.
Each question in PoK requires just one API call, used solely in the Plan of Knowledge stage; retrieval and reasoning are fully handled by our fine-tuned open-source model.

Prompt length further contributes to efficiency. PoK uses a fixed template (134 tokens), the question itself (13.01 tokens on average), and 20 retrieved facts (67.2 tokens each). In contrast, RTQA constructs custom prompts for each question type and retrieves up to 50 facts, leading to much longer inputs and slower inference.

\subsection{Case Study} \label{sec:case}

We illustrate PoK's and GPT-4o's reasoning behaviors on overlap-type 3-hop and after/first-type 2-hop questions in \tbref{tab:3hop} and \tbref{tab:2hop}, respectively.
From the cases, we observe that for complex temporal questions, the Plan of Knowledge framework enables step-by-step reasoning rather than attempting to produce the final answer in a single step.
This decomposition allows the model to consider intermediate facts and temporal constraints explicitly, which improves the accuracy of the final answers. 
In contrast, GPT-4o tends to generate a list of responses that are partially correct, incomplete, or even irrelevant.
Furthermore, after fine-tuning, PoK's outputs align more closely with standard answer formats.  GPT-4o, however, often produces semantically correct answers that fail to match the required format, which leads to incorrect evaluation scores. For example, when the ground-truth answer is "2012-05" for a question such as “in which month…," GPT-4o frequently outputs “May," which, although semantically valid, is considered format-inconsistent and thus counted as incorrect during evaluation.

\section{Conclusion and Future Work} \label{conclusion}

In this work, we address two challenges faced by LLMs in handling temporal questions and propose PoK, a Plan of Knowledge framework enhanced with a time-aware retriever. The framework decomposes complex temporal questions into structured sub-objectives and leverages a fine-tuned Temporal Knowledge Store (TKS) for retrieving semantically and temporally aligned facts. In addition, task-specific temporal prompts are incorporated to strengthen temporal awareness during reasoning.
Extensive experiments on four benchmark TKGQA datasets verify its effectiveness, achieving relative performance gains of up to 56.0\% over existing baselines.

Although our approach achieves substantial improvements, the retrieval of complex temporal facts remains a challenge. Future research should explore methods for retrieving more precise and effective temporal information, especially in the context of incomplete TKGQA, where missing temporal facts hinder reliable reasoning.
Additionally, controlling the answer format during the generation process of LLMs without fine-tuning is difficult, as discussed in Section \ref{sec:case}. Thus, standardizing the answer formats of LLMs or developing a more reasonable evaluation method is another important future task.

% \section*{Acknowledgments}
% This is a simple paragraph thanking those who supported your work.

\bibliographystyle{IEEEtran}
\bibliography{bare_jrnl_new_sample4}   % custom.bib 不加 .bib

\vfill

\end{document}